\documentclass{article}
\usepackage[compatibility=false]{caption}
\usepackage{spconf,amsmath,graphicx}
\usepackage{tabularx}
\usepackage{booktabs}
\usepackage{multirow}
\usepackage{amsmath}
\usepackage{amssymb}
\usepackage{lscape}
\usepackage{media9}
\usepackage{caption}
\usepackage{hyperref}
\usepackage{array}
\usepackage{float}
\usepackage{graphicx,scalerel}
\usepackage{comment}
\usepackage[most]{tcolorbox}
\usepackage{enumitem}
\usepackage{hhline}
\usepackage{makecell}

\definecolor{vdbest}{rgb}{0.96, 0.57, 0.58}
\definecolor{vdsecond}{rgb}{0.98, 0.78, 0.57}
\definecolor{vdthird}{rgb}{1.0, 1.0, 0.56}
\def \y  {\textcolor{black}{\checkmark}}
\title{Veta-GS: View-dependent deformable 3D Gaussian Splatting \\ for thermal infrared Novel-view Synthesis}

\name{Myeongseok Nam$^{1*}$\thanks{Equal contribution.}, Wongi Park$^{1*}$\footnotemark[1], Minsol Kim$^{2}$, Hyejin Hur$^{3}$, and Soomok Lee$^1$\sthanks{Corresponding author.}\vspace{-1em}}
\address{$^{1}$Ajou University \quad $^{2}$Sejong University \quad $^{3}$Korea University \\
        \normalsize{\texttt{\{ncmssh0313$^1$, soomoklee$^5$\}@ajou.ac.kr}}  \\ 
        {\textbf{\textcolor{magenta}{\url{https://nbril0313.github.io/Veta-GS/}}}}}
  
\begin{document}

\maketitle
\begin{abstract}
Recently, 3D Gaussian Splatting (3D-GS) based on Thermal Infrared (TIR)  imaging has gained attention in novel-view synthesis, showing real-time rendering. However, novel-view synthesis with thermal infrared images suffers from transmission effects, emissivity, and low resolution, leading to floaters and blur effects in rendered images. To address these problems, we introduce \textbf{Veta-GS}, which leverages a view-dependent deformation field and a Thermal Feature Extractor (TFE) to precisely capture subtle thermal variations and maintain robustness. Specifically, we design view-dependent deformation field that leverages camera position and viewing direction, which capture thermal variations. Furthermore, we introduce the Thermal Feature Extractor (TFE) and MonoSSIM loss, which consider appearance, edge, and frequency to maintain robustness. Extensive experiments on the TI-NSD benchmark show that our method achieves better performance over existing methods.
\end{abstract}

\begin{figure*}[!t]
    \centering
    \includegraphics[width=\textwidth,height=0.27\textheight]{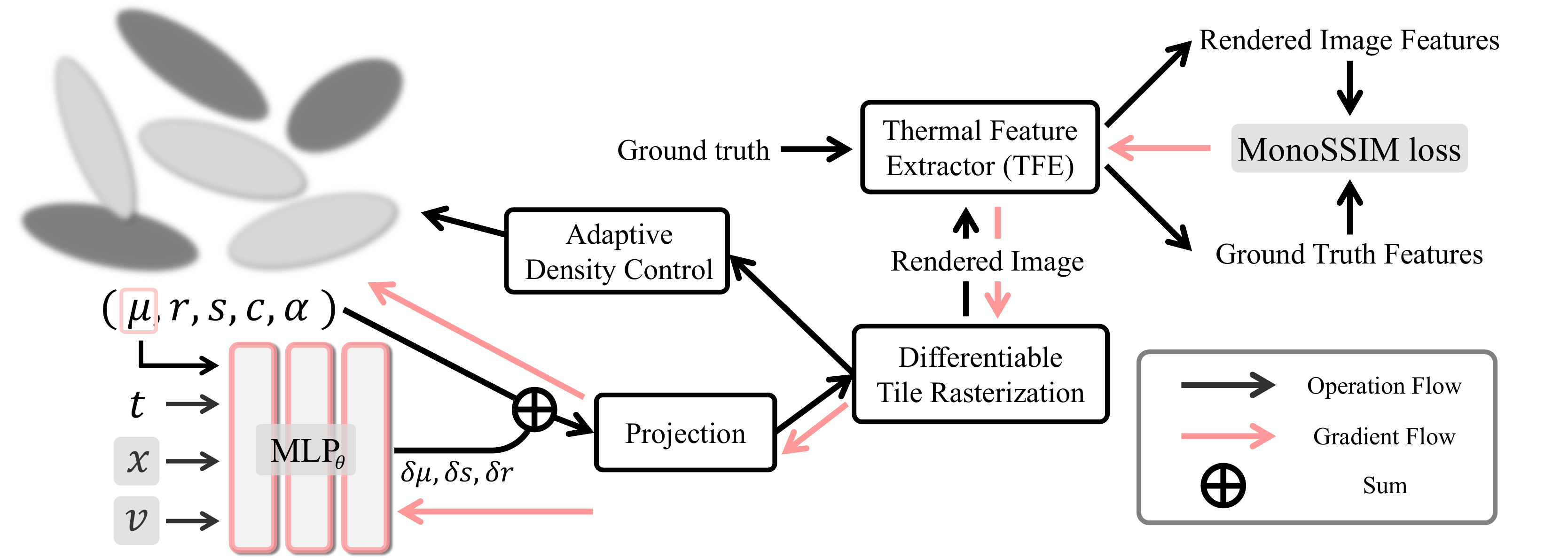}
    \caption{Overview of Veta-GS pipeline. We utilize camera position $x$, viewing direction $v$, time $t$, and 3D Gaussian’s position \(\mu\) with positional encoding as input to deform 3D gaussians, by obtaining the offset $(\delta \mu,\;\delta r,\;\delta s)$. Further we introduce Thermal Feature Extractor (TFE) and MonoSSIM loss which focus on appearance, edge, and frequency to show robustness rendering.}
\end{figure*}

\section{Introduction}
\indent Novel-view synthesis has been widely studied for generating photorealistic images from arbitrary viewpoints using only a sparse set of captured 2D images. This fundamental task in computer vision and graphics underpins many applications, including augmented/virtual reality (AR/VR) \cite{xia2024video2game}, autonomous driving \cite{zhou2024drivinggaussian, yang2023emernerf}, and 3D content creation \cite{liu2024one}. While implicit neural representation methods such as Neural Radiance Fields (NeRF) have demonstrated remarkable visual fidelity, their slow rendering speeds can limit the application in real-time rendering. \\
\indent To mitigate this issue, 3D Gaussian Splatting (3D-GS) \cite{kerbl20233d, sabour2024spotlesssplats} has emerged as an explicit 3D representation that shows real-time rendering and high quality 3D novel-view synthesis. While 3D-GS \cite{kerbl20233d, sabour2024spotlesssplats} works based on visible light images are widely studied, approaches based on TIR images is still under-explored. Unlike visible light imaging, TIR imaging relies on thermal radiation and is highly sensitive to fluctuations due to transmission effects, emissivity, and low resolution. Consequently, directly applying novel-view synthesis with thermal infrared images results in noise, floaters, and blur artifacts. \\
\indent To tackle these problems, several methods \cite{lin2024thermalnerf, lu2024thermalgaussian, chen2024thermal3d, ye2024thermal} have begun to emerge in the thermal infrared novel-view synthesis domain. ThermalNeRF \cite{lin2024thermalnerf} utilizes a multi-modal training strategy with RGB images to address floaters and blur effects. However, the multi-modal training strategy is limited by the requirement to collect both RGB and TIR images. Recently, concurrent works \cite{ye2024thermal, chen2024thermal3d} have utilized only TIR images. Thermal-NeRF \cite{ye2024thermal} exploits structural information to enhance robustness. Thermal3D-GS \cite{chen2024thermal3d} considers physical phenomena to address transmission effects and emissivity. \\
\indent Although existing methods demonstrate impressive results, they still suffer from two limitations. Firstly, it struggles to handle the thermal variations that occur with changes in viewpoint, including the transmission effects reflected from the surrounding environment. Secondly, TIR images inherently have low resolution and often suffer from noise, which occurs due to TIR imaging properties. To address these problems, we propose Veta-GS, which leverages a view-dependent deformation field and a Thermal Feature Extractor (TFE) to capture subtle thermal variations and maintain robustness. \\
\indent We design a view-dependent deformation field and a frustum-based masking strategy that leverage the camera position and viewing direction to capture thermal variations and accelerate training speed. Furthermore, we introduce a Thermal Feature Extractor (TFE) and MonoSSIM loss, which consider appearance, edge, and frequency and to maintain robustness. Extensive experimental results on the TI-NSD benchmark show that Veta-GS demonstrates better performance compared to existing methods. In summary, our main contributions are as follows: \\
• We propose Veta-GS, which utilize view-dependent deformation field and a frustum-based masking strategy that leverage the camera position and viewing direction to precisely capture thermal variations and boost rendering speed. \\
\noindent • We introduce Thermal Feature Extractor (TFE) and MonoSSIM loss to maintain robustness in rendered images by considering appearance, edge, and frequency. \\
\noindent • Extensive experiments conducted on TI-NSD benchmark show that our method improves the rendering quality compared to existing methods across indoor and outdoor scenes.
\vspace{-1.4em}
\section{Related works}
\noindent\textbf{Novel-view synthesis.} Novel-view synthesis has achieved groundbreaking results using sparse sets of captured 2D images. Traditional approaches rely on Structure-from-Motion (SfM) \cite{schonberger2016structure} and Multi-View Stereo (MVS) \cite{chaurasia2013depth} for scene reconstruction. While these methods demonstrate favorable performance, they require high computational cost. To address this, Neural Radiance Fields (NeRF) \cite{mildenhall2020nerf} have emerged as a prominent solution by learning implicit radiance fields. However, these methods still suffer from the long training time associated with volumetric rendering. Recently, 3D Gaussian Splatting (3D-GS) \cite{kerbl20233d} has gained significant attention as an innovative method for novel-view synthesis. Despite these rapid advancements, prior methods have predominantly focused on high-quality RGB images, with limited exploration of TIR images. In this work, we tackle this gap by addressing TIR image reconstruction. \\
\noindent\textbf{Thermal Infrared (TIR) imaging.} 
Recently, following the success of novel-view synthesis \cite{mildenhall2020nerf, yu2021plenoxels, muller2022instant}  based on RGB images, novel-view synthesis using TIR images has gathered attention in 3D reconstruction. TIR imaging has inherent limitations such as transmission effects, emissivity, and low resolution. Previous works have used TIR images along with RGB images to address these problems. However, the necessity of RGB images limits the applicability in a wide range of fields. To mitigate this issue, Thermal3D-GS \cite{chen2024thermal3d} using only TIR images have emerged either by modeling physics-induced phenomena or focusing on structural information. Even though above methods show impressive results, these methods don't consider thermal variation dependent on the viewpoint. In this work, we apply deformation in a view-dependent way to capture subtle thermal variations that vary over viewpoints. 
\vspace{-1.2em}
\section{Preliminaries}
\label{sec:pagestyle}
\noindent
\textbf{3D Gaussian Splatting (3D-GS).}
In 3D-GS \cite{kerbl20233d}, a scene is represented as an explicit set of learnable primitives 
\(\{\mathcal{G}_0,\,\mathcal{G}_1,\,\dots,\,\mathcal{G}_N\}\).
Each primitive \(\mathcal{G}_i\) defined by a position \(\mu_i \in \mathbb{R}^{3}\), a covariance matrix 
\(\boldsymbol{\Sigma}_i \in \mathbb{R}^{3 \times 3}\), opacity \(\boldsymbol{o_i}\in\mathbb{R}\), spherical harmonics coefficients \(SH_i\in\mathbb{R}^L\). 
To optimize the parameters of 3D Gaussians, we need to render these 3D Gaussians to a 2D image. The 3D Gaussians's position \(\mu_i\) and covariance matrix \(\boldsymbol{\Sigma}_i\) are transformed using a viewing matrix \(\boldsymbol{W}\) and an affine approximation \(\boldsymbol{J}\). The resulting 2D parameters, position \(\hat{\mu}_i \in \mathbb{R}^2\) and covariance \(\hat{\boldsymbol{\Sigma}}_i \in \mathbb{R}^{2 \times 2}\), define the kernel of each projected Gaussian in the 2D image plane. The kernel at a pixel coordinate \(\boldsymbol{p}\in\mathbb{R}^2\) is given by:
\begin{equation}
    \hat{G}_i(\boldsymbol{p}) = \exp\Bigl(-\tfrac{1}{2}\bigl(\boldsymbol{p}-\hat{\mu}_i\bigr)^\top\hat{\boldsymbol{\Sigma}}_i^{-1}\bigl(\boldsymbol{p}-\hat{\mu}_i\bigr)\Bigr).
    \label{eq:2d-density-prelim}
\end{equation} The pixel color \(C(\boldsymbol{p})\) is computed by sorting the Gaussians according to depth and performing alpha blending:
\begin{equation}
  C(\boldsymbol{p})
  \;=\;
  \sum_{i=1}^N \,c_i\,\alpha_i\,\prod_{j=1}^{i-1}\,\bigl(1 - \alpha_j\bigr),
  \quad
  \alpha_i = \sigma\bigl(\boldsymbol{o_i}\bigr)\,\hat{G}_i(\boldsymbol{p}),
  \label{eq:gs-alpha-prelim}
\end{equation}
where \(\sigma(\cdot)\) denotes the sigmoid function, \(c_i\) is the view-dependent color derived from \(SH_i\), and \(N\) is the number of Gaussians affecting the pixel \(\boldsymbol{p}\).
The reconstruction loss is optimized through D-SSIM\cite{wang2004image} term and \(\mathcal{L}_{1}\):
\begin{equation}
    \mathcal{L}_{\text{3DGS}} = (1 - \lambda)\mathcal{L}_{1} + \lambda\mathcal{L}_{\text{D-SSIM}},
\end{equation} where $\lambda$ is 0.2. This formulation ensures a balance between pixel-wise accuracy and perceptual similarity.
\begin{figure}
    \centering
    \includegraphics[width=\linewidth]{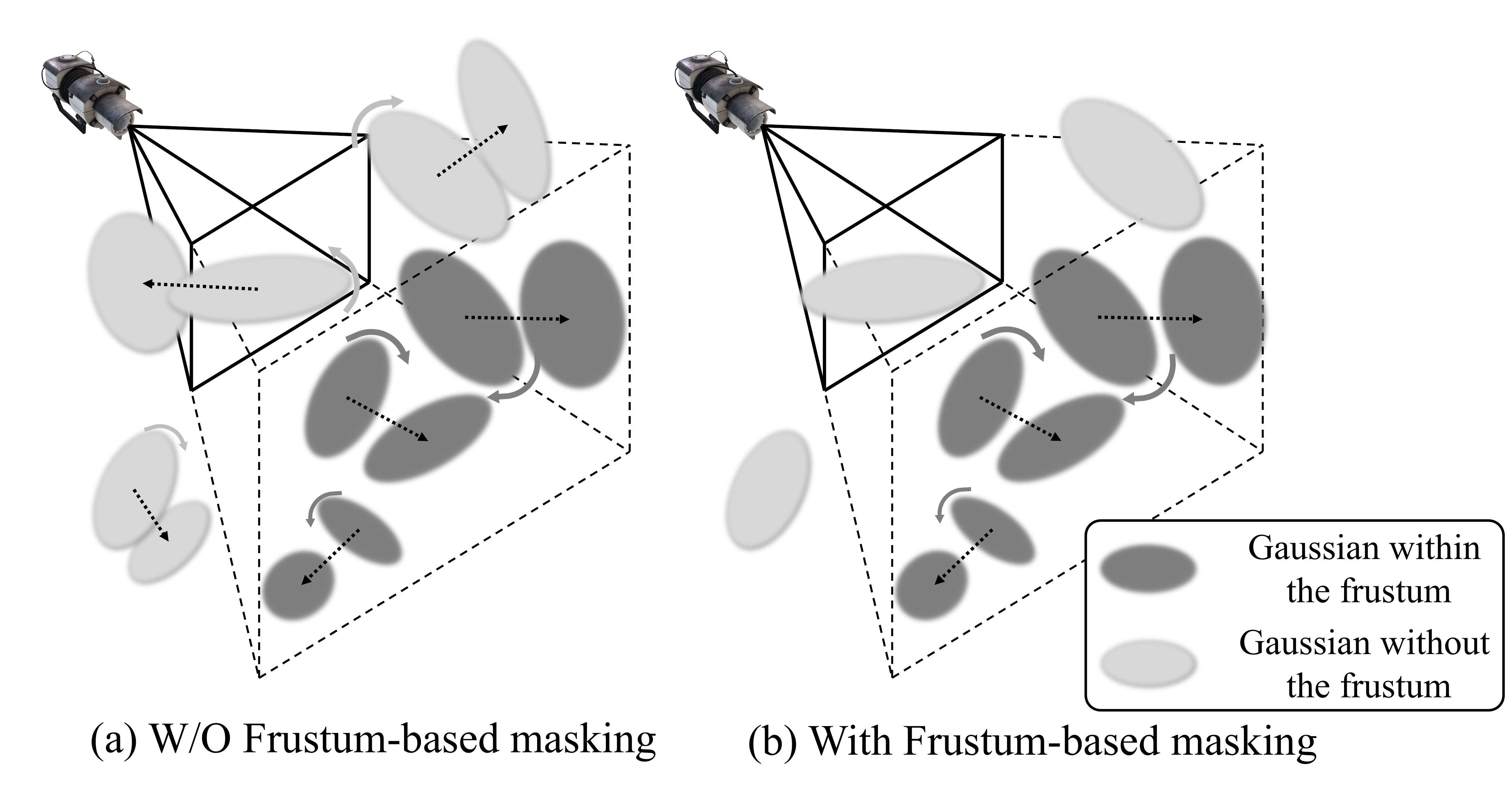}
    \caption{Frustum-based masking. (a) Existing method deforms all 3D gaussians. On the other hand, (b) we identify the 3D Gaussians bounded by the frustum, then selectively apply deformation only to those 3D Gaussians, thereby accelerate training speed.}
    \label{fig:frustum}
\vspace{-1em}
\end{figure}
\begin{figure*}[!t]
\centering
\includegraphics[width=\textwidth]{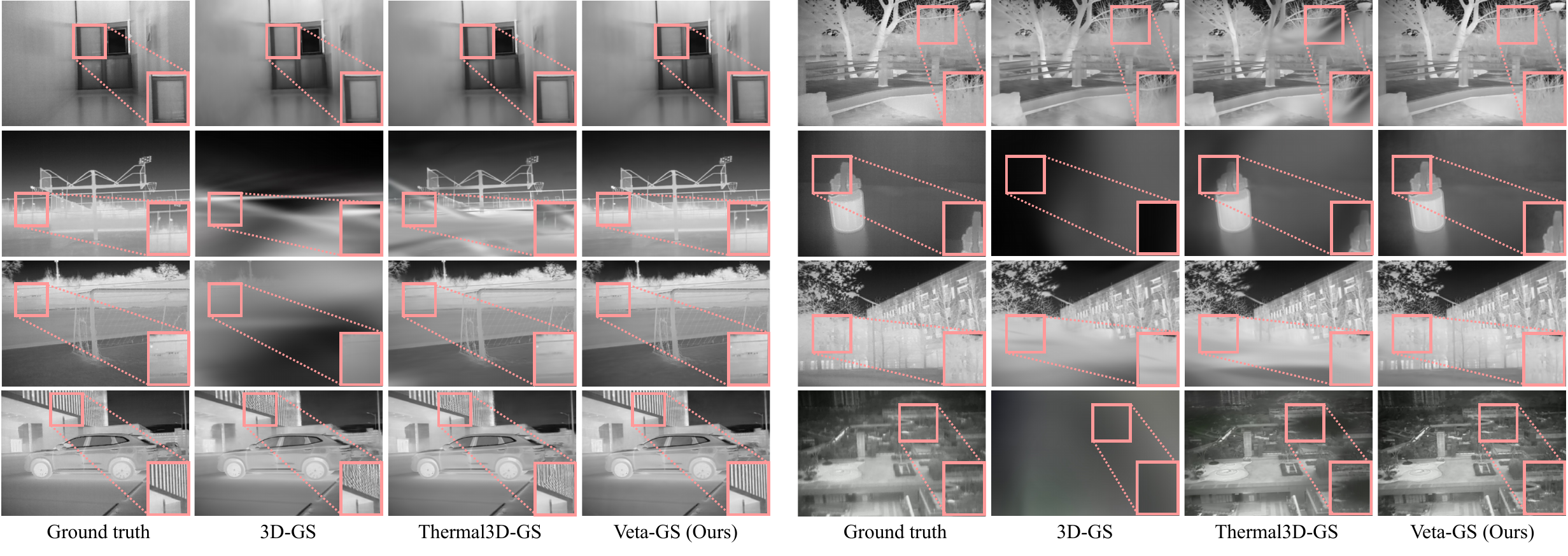}
\caption{Visualization of experiments on TI-NSD dataset including indoor scenes and outdoor scenes. Experiment results demonstrate that Veta-GS shows high-quality 3D rendering across small-scale scenes to large-scale scenes, while 3D-GS\cite{kerbl20233d} and Thermal3D-GS\cite{chen2024thermal3d} struggle with floater artifacts and thermal variations.}
\label{fig:sota}
\end{figure*}
\vspace{-0.5em}
\section{Methods}
\noindent 
\noindent \textbf{Deformable 3D Gaussians Fields.} As intuitive solution to model dynamic scene, existing method \cite{yang2024deformable} utilizes Deformable MLP \(\mathcal{F}_{\theta}\) module which involve time $t$ with position encoding and 3D Gaussian’s position $\mu$ to make deformable space as such that:
\begin{equation}
(\delta \mu,\;\delta r,\;\delta s)
\;=\;
\mathcal{F}_{\theta}\bigl(\gamma\!\bigl(\mathrm{sg}(\mu)\bigr),\;\gamma(t)\bigr),
\label{eq:DeformableGaussianExisting}
\end{equation} where \(\mathrm{sg}(\cdot)\) indicates a stop-gradient operation and \(\gamma(\cdot)\) is a positional encoding. The resulting parameters \(\bigl(\mu+\delta \mu,\;r+\delta r,\;s+\delta s\bigr)\) are then used to represent dynamic objects through time. However, directly applying the aforementioned method in novel-view synthesis with TIR images fails to capture the thermal variations corresponding to different viewpoints. To address
this issue, we utilize the deformation network to incorporate view-dependent terms. Specifically, we augment the Deformable MLP \(\mathcal{F}_{\theta}\) input with the camera position $x$ and viewing direction $v$ as such that:
\begin{equation}
(\delta \mu,\;\delta r,\;\delta s)
\;=\;
\mathcal{F}_{\theta}\Bigl(
  \gamma\!\bigl(\mathrm{sg}(\mu)\bigr),\;
  \gamma(t),\;
  x,\;
  v
\Bigr).
\label{eq:view-deform}
\end{equation} Additionally, we observe that previous deformation methods \cite{yang2024deformable,lu20243d} deform all Gaussians to tackle scenarios where extensive motion is expected. However, the fundamental goal of novel-view synthesis based on TIR images is to handle subtle thermal variations. To tackle this issue, we introduce a frustum-based masking strategy that identifies the Gaussians bounded by the frustum and then selectively applies deformation only to those Gaussians, as shown in Fig. \ref{fig:frustum}. To the best of our knowledge, this is the first method to leverage the deformation of Gaussians within the frustum, thereby accelerating rendering speed and quality. \\
\noindent \textbf{Thermal Feature Extractor and MonoSSIM loss.} Novel-view synthesis with TIR imaging suffers from floaters and blurring effects due to transmission effects and low resolution. Inspired by SSIM \cite{wang2004image}, which compares the luminance, contrast, and structure of images to measure their similarity, we adapt these properties to align with the characteristics of TIR images. We compare the rendered image and GT image using the TFE module from three perspectives: appearance, edge, and frequency. Each component is processed by a distinct CNN module as follow:

\begin{tcolorbox}[colframe=gray!30, colback=white, coltitle=black, arc=10pt, boxrule=0.8mm]
    \textbf{• Appearance $\Phi_{App}$}: Extracts appearance features from raw TIR images using a CNN, focusing on visual texture and surface details. \\
    \textbf{• Edge $\Phi_{Edg}$}: Extracts edge features from edge-enhanced TIR images (e.g., via Canny) using a CNN, emphasizing object boundaries and contours.\\
    \textbf{• Frequency $\Phi_{Frq}$}: Extracts frequency features from fourier-transformed TIR images using a CNN, capturing high-frequency details in texture.
\end{tcolorbox} \noindent  Let \(\mathbf{I}_{\mathrm{pred}}\) and \(\mathbf{I}_{\mathrm{gt}}\) be the rendered and ground-truth images, respectively. For each aspect \(r \in \{App,\,Edg,\,Frq\}\), we have a distinct CNN module \(\Phi_r(\cdot)\). We obtain:
\begin{equation}
f_{\mathrm{pred}}^{(r)} \;=\; \Phi_{r}\!\bigl(\mathbf{I}_{\mathrm{pred}}\bigr),
\quad
f_{\mathrm{gt}}^{(r)} \;=\; \Phi_{r}\!\bigl(\mathbf{I}_{\mathrm{gt}}\bigr).
\end{equation} 
We compute the MonoSSIM loss from the features as follows:
\begin{equation}
\mathcal{L}_{\text{Mono}} = (1-\mathcal{L}_{App})\cdot(1-\mathcal{L}_{Edg})\cdot(1-\mathcal{L}_{Frq}).
\end{equation} We calculate the loss by utilizing the features from each aspect through CNN $\Phi_{r}$ to compute the cosine similarity loss. Specifically, we change three components as appearance, edge, and frequency features. Inspired by NeRF-on-the-go \cite{ren2024nerf}, instead of using D-SSIM’s original formulation \cite{wang2004image},  \(\mathcal{L}_{\text{D-SSIM}} = 1 - \mathcal{L}_{\text{L}} \cdot \mathcal{L}_{\text{C}} \cdot \mathcal{L}_{\text{S}}\,\), we utilize an updated formulation to prevent one aspect being underestimated when another aspect is near zero. This ensures a balanced contribution from all three aspects, effectively maintaining robustness in thermal infrared novel-view synthesis.

\begin{table*}[!t]
\centering
\renewcommand{\arraystretch}{0.9}
\resizebox{\textwidth}{!}{
\begin{tabular}{ccccccccccccc}
\toprule
\multirow{2}{*}{Method} & \multicolumn{3}{c}{Indoor} & \multicolumn{3}{c}{Outdoor} & \multicolumn{3}{c}{UAV} & \multicolumn{3}{c}{Average} \\
          & PSNR$\uparrow$    & SSIM$\uparrow$    & LPIPS$\downarrow$  & PSNR$\uparrow$    & SSIM$\uparrow$    & LPIPS$\downarrow$   & PSNR$\uparrow$   & SSIM$\uparrow$   & LPIPS$\downarrow$ & PSNR$\uparrow$    & SSIM$\uparrow$    & LPIPS$\downarrow$   \\ 
\midrule
Plenoxels\cite{yu2021plenoxels}        & 22.13   & 0.867   & 0.385  & 22.15   & 0.768   & 0.433   & 25.56  & 0.780  & 0.351 & 23.28   & 0.805   & 0.390   \\
INGP-Base\cite{muller2022instant}        & 26.99   & 0.916   & 0.291  & 26.00   & 0.837   & 0.302   & 20.86  & 0.681  & 0.404 & 24.62   & 0.811   & 0.332   \\
INGP-Big\cite{muller2022instant}         & 27.46   & 0.918   & 0.289  & 26.45   & 0.839   & 0.292   & 20.82  & 0.680  & 0.387 & 24.91   & 0.812   & 0.323   \\
3D-GS\cite{kerbl20233d}        & 32.98   & 0.953   & 0.259  & 28.89   & 0.904   & 0.227   & 34.51  & 0.953  & 0.119 & 32.01   & 0.936   & 0.206   \\
Thermal3D-GS\cite{chen2024thermal3d}      & 36.01   & 0.962   & 0.252  & 32.60   & 0.942   & 0.187   & 36.74  & 0.962  & 0.112 & 35.04   & 0.955   & 0.187   \\
Veta-GS \textbf{(Ours)} & \textbf{37.07}       & \textbf{0.963}   & \textbf{0.240} & \textbf{33.78}     & \textbf{0.947}       & \textbf{0.168}  & \textbf{37.05} & \textbf{0.963} & \textbf{0.100}     & \textbf{35.97}       & \textbf{0.958}       & \textbf{0.169}       \\
\bottomrule
\end{tabular}}
\caption{Quantitative results with state-of-the-art methods on the TI-NSD datasets across small-scale scenes to large-scale scenes. The best numbers within each block are bolded.}
\label{Tab:SoTA}
\end {table*}
\noindent
\textbf{Optimization.} During the training of Veta-GS, we follow existing densification strategy \cite{kerbl20233d}. Since the rendered images are initially incomplete, we apply TFE and MonoSSIM loss after 20K iterations to ensure stable training. Before 20K iteration, the total loss compute as follow:
\begin{equation}
\mathcal{L}_{\text{total}} = (1 - \lambda)\,\mathcal{L}_{1} + \lambda\,\mathcal{L}_{\text{D-SSIM}},
\label{eq:loss_before_20000}
\end{equation} where $\lambda$ is 0.2. After 20K iterations, we incorporate the MonoSSIM loss, and the final loss is computed as follows:
\begin{equation}
\mathcal{L}_{\text{total}} = (1 - \lambda_{\text{Mono}} - \lambda)\,\mathcal{L}_{1} 
+ \lambda_{\text{Mono}}\,\mathcal{L}_{\text{Mono}} 
+ \lambda\,\mathcal{L}_{\text{D-SSIM}},
\label{eq:loss_after_20000}
\end{equation}
where $\lambda_{\text{Mono}}$ is also 0.2. This optimization strategy stabilizes densification and simultaneously improves rendering quality.

\begin{table}[!t]
\centering
\renewcommand{\arraystretch}{0.7} 
\begin{tabular}{cccccc}
\toprule
methods & PSNR$\uparrow$ & SSIM$\uparrow$ & LPIPS$\downarrow$ & Times$\downarrow$ &\\ 
\midrule
Full + w/o FM   & 33.36    & 0.954    & 0.162              & 127min     \\
Full            & 33.49    & 0.954    & 0.162              & 85min     \\
\bottomrule
\end{tabular}
\caption{Comparison of performance and training time with and without frustum-based masking.}
\label{Tab:FM}
\vspace{-1.5em}
\end{table}
\section{Experiment} 
\vspace{-0.5em}
\noindent\textbf{Dataset and Metrics.} We conduct our experiments on the Thermal Infrared Novel-view Synthesis Dataset (TI-NSD) across small scenes to large-scale scenes, demonstrating our proposed method. We report PSNR, SSIM \cite{wang2004image}, and LPIPS \cite{zhang2018unreasonable} scores to evaluate the performance quantitatively.  \\
\noindent\textbf{Baseline and Implementation details.} Our method is built upon Deformable 3D Gaussians \cite{yang2024deformable}, and we follow default hyperparameter setup for training our method. We design depth of layer in TFE module as $D = 2$. We increased the densification iterations from 15K to 20K and experimentally determined the densification threshold to avoid exceeding GPU memory limitations. The Adam optimizer was utilized, and a cosine annealing strategy was employed to gradually decrease the learning rate to $1.6 \times 10^{-6}$.  We used a single RTX GPU(24GB) for all experiments. \\
\noindent\textbf{Comparision with existing methods.} We compare Veta-GS with previous methods including Plenoxels \cite{yu2021plenoxels}, InstantNGP (INGP) \cite{muller2022instant}, 3D-GS\cite{kerbl20233d}, and Thermal3D-GS \cite{chen2024thermal3d} in the Tab. \ref{Tab:SoTA}. Among them, Plenoxels \cite{yu2021plenoxels} and InstantNGP\cite{muller2022instant} could not capture thermal variations. Furthermore, 3D-GS\cite{kerbl20233d} and Thermal3D-GS \cite{chen2024thermal3d} often suffer from capturing high-frequency details due to transmission effects and emissivity as shown in Fig. \ref{fig:sota}. However, our proposed method Veta-GS exploits a view-dependent deformation field to precisely capture thermal variations. In addition, we introduce the TFE module and MonoSSIM loss, which consider appearance, edge, and frequency. Finally, we demonstrate that Veta-GS achieves high-quality rendering and alleviates floaters and blur artifacts across indoor, outdoor, and UAV scenarios on the TI-NSD benchmark.\\
\vspace{-2.3em}
\section{Ablation Study}
\begin{figure}[!t]
    \centering
    \includegraphics[width=\linewidth]{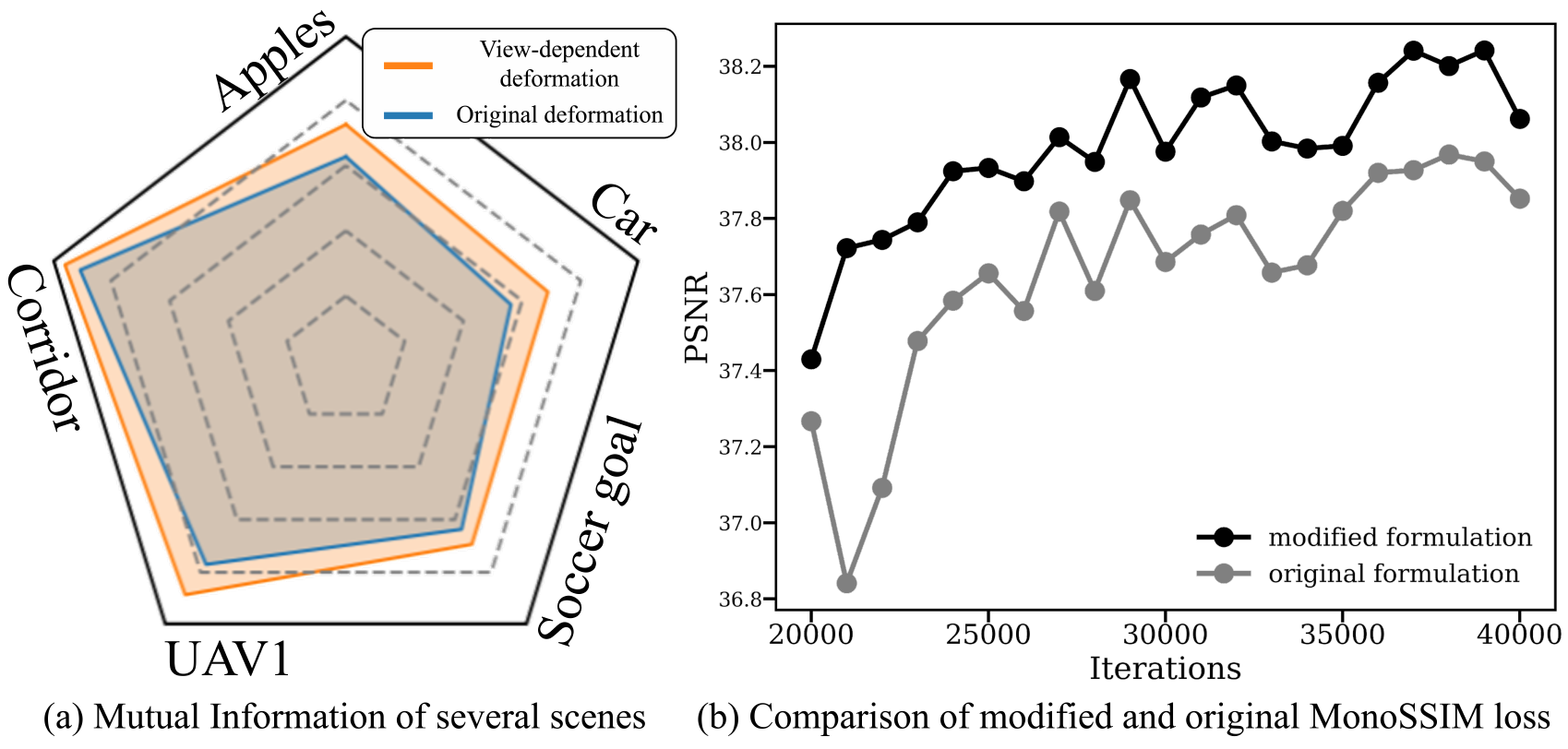}
    \caption{(a) Comparison of Mutual Information (MI) between view-dependent and time embedding on several scenes. (b) Comparison of training convergence curve between the modified and original formulation of $\mathcal{L}_{\text{Mono}}$.}
    \label{fig:MI}
\end{figure}
\begin{table}[!t]
\centering
\renewcommand{\arraystretch}{0.9} 
\resizebox{\columnwidth}{!}{
\begin{tabular}{cccccc}
\toprule
VDF & FM & TFE &  PSNR$\uparrow$ & SSIM$\uparrow$ & LPIPS$\downarrow$ \\ 
\midrule
    &     &     & 32.14          & 0.937          & 0.202 \\
\y  &     & \y  & 35.10          & 0.954          & 0.176 \\
\y  & \y  &     & 36.01          & 0.955          & 0.174     \\
\y  & \y  & \y  & 35.97          & 0.958          & 0.169     \\
\bottomrule
\end{tabular}}
\caption{Each module analysis on the Veta-GS. VDF and FM denote view-dependent deformable field and Frustum-based masking. TFE denotes the two combined Thermal Feature Extractor and MonoSSIM loss.}
\label{Tab:EachModule}
\end{table}
\noindent\textbf{View-dependent deformation.} We analyze efficiency to validate effectiveness of proposed methods on bicycle scene, we conduct experiments in Tab. \ref{Tab:FM}. We observed that the Frustum-based masking strategy significantly contributes to accelerating training speed. Furthermore, we measure the mutual information as shown in Fig. \ref{fig:MI}-(a), comparing the case where only the time embedding is used as input and the case where the viewing direction and camera pose are input into the deformable MLP $\mathcal{F}_{\theta}$. This proves that the time embedding as input is independent, whereas the camera pose and viewing direction as input are highly correlated. Therefore, by considering the transformation of the camera pose and viewing direction, it is feasible to capture the thermal variations across different views. \\

\begin{table}[!t]
\renewcommand{\arraystretch}{0.84} 
\centering
\resizebox{\columnwidth}{!}{
\begin{tabular}{cccccc}
\toprule
$\mathcal{L}_{App}$ & $\mathcal{L}_{Edg}$ & $\mathcal{L}_{Frq}$ &  PSNR$\uparrow$ & SSIM$\uparrow$ & LPIPS$\downarrow$ \\ 
\midrule
    &     &     & 35.74          & 0.952          & 0.282 \\
\y  &     &     & 36.35          & 0.961          & 0.270 \\
    & \y  &     & 36.57          & 0.963          & 0.266 \\
    &     & \y  & 37.54          & 0.967          & 0.260  \\
\y  & \y  &     & 37.14          & 0.966          & 0.261     \\
\y  & \y  & \y  & 37.58          & 0.958          & 0.169 \\
\bottomrule 
\end{tabular}}    
\caption{Comparison of combinations of various loss forms.}
\label{tab:mono_module}
\vspace{-1em}
\end{table}
\noindent \textbf{MonoSSIM Loss.} In Fig. \ref{fig:MI}-(b), we visualize the convergence curve, demonstrating that the performance improvement converges progressively. Furthermore, to assess the impact of modified $\mathcal{L}_{\text{Mono}}$, we conduct extensive experiments in Tab. \ref{tab:lambda}. \\
\begin{table}[!t]
    \centering
    \footnotesize
    \renewcommand{\arraystretch}{0.8} 
    \begin{tabular}{ccccccc}
    \toprule
        \multicolumn{7}{c}{Hyperparameter $\lambda_\text{Mono}$}          \\
        \midrule
        0.05 & 0.10 & 0.15 & 0.20 & 0.25 & 0.30 & 0.35  \\
        \midrule                 
        34.66 & 34.85 & 34.36 & \textbf{35.26} & 34.04 & 34.85 & 34.29  \\ \bottomrule
    \end{tabular}
    \caption{
    Experimental result on the hyperparameter $\lambda_\text{Mono}$ setup to assess the impact $\mathcal{L}_{\text{Mono}}$.}
    \label{tab:lambda}
\end{table}
\begin{figure}
    \centering
    \includegraphics[width=\linewidth]{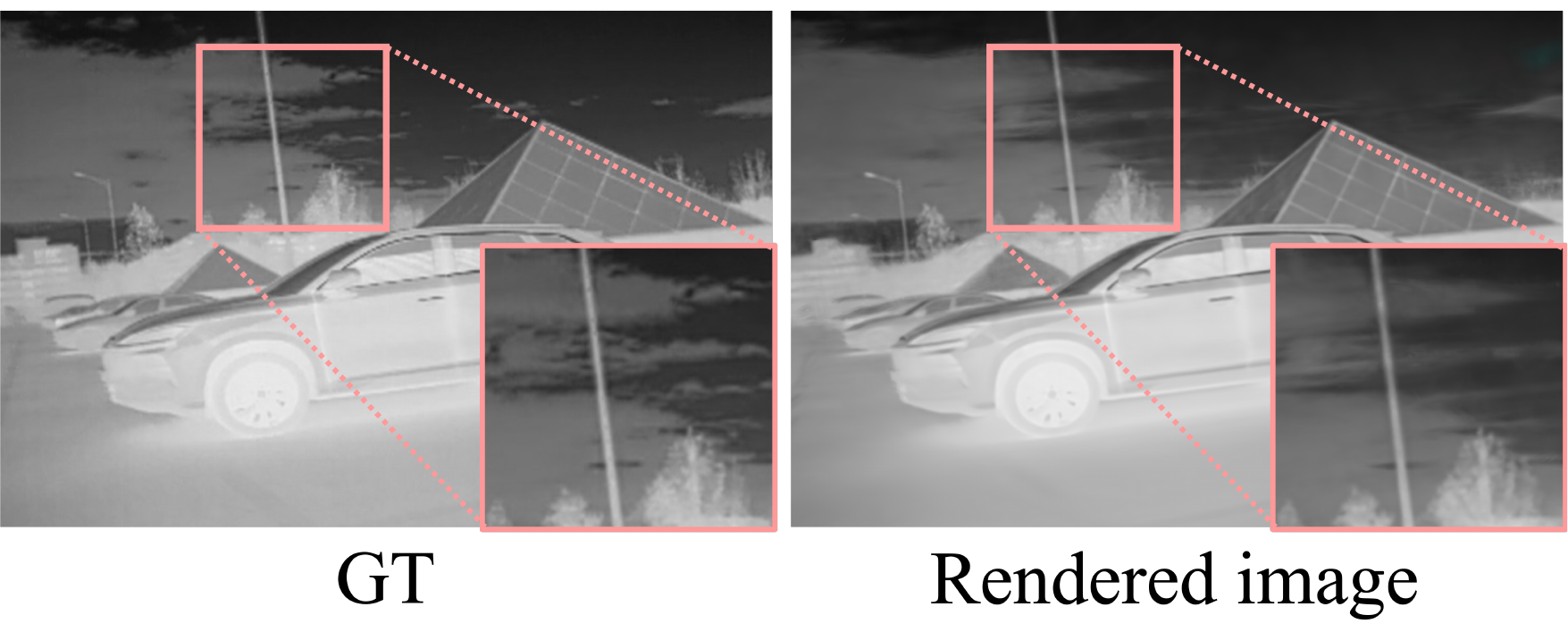}
    \caption{Failure cases on textures with significant variations.}
    \label{fig:limitation}
\vspace{-1em}
\end{figure}
\noindent \hspace{-0.25em}\textbf{Component Analysis.} To validate effectiveness of proposed methods, we conduct experiments in Tab. \ref{Tab:EachModule}. We observed that the performance improved as each module was progressively added, and confirmed that the $\mathcal{L}_{Frq}$ significantly contributed to the quantitative performance enhancement in Tab. \ref{tab:mono_module}.
\section{Limitation and Conclusion}
\vspace{-1em}
\noindent \textbf{Limitation.} Although our method effectively preserves high-frequency details through the MonoSSIM loss, it still struggles to capture textures with substantial variations, such as clouds and leaves as depicted in Fig.\ref{fig:limitation}. Another limitation is the computational cost of TFE. While TFE is lightwweight with only two layers, the overhead is not trivial due to the usage of three separate TFEs independently handling appearance, edge, frequency features. \\
\noindent \textbf{Conclusion.} This paper presents Veta-GS, which leverages a view-dependent deformation field and a Thermal Feature Extractor (TFE) to precisely capture subtle thermal variations while maintaining robustness. Additionally, we design a frustum-based masking strategy to accelerate rendering speed. Moreover, we propose a Thermal Feature Extractor (TFE) and MonoSSIM loss, which consider appearance, edge, and frequency to enhance robustness. Experimental results demonstrate that our method achieves significant performance. In future work, we believe it would be beneficial to investigate dynamic Thermal Infrared (TIR) imaging and expand to address complex dynamic environments. 

\section{Acknowledgement.} 
\vspace{-1em}
\noindent This work was supported in 2024 by Korea National Police Agency(KNPA) under the project "Development of autonomous driving patrol service for active prevention and response to traffic accidents"(RS-2024-00403630).

\vfill
\newpage

\bibliographystyle{IEEEbib}
\bibliography{strings}

\begin{thebibliography}{10}

\bibitem{xia2024video2game}
Hongchi Xia, Zhi-Hao Lin, Wei-Chiu Ma, and Shenlong Wang,
\newblock ``Video2game: Real-time interactive realistic and browser-compatible environment from a single video,''
\newblock in {\em IEEE Conference on Computer Vision and Pattern Recognition}, 2024, pp. 4578--4588.

\bibitem{zhou2024drivinggaussian}
Xiaoyu Zhou, Zhiwei Lin, Xiaojun Shan, Yongtao Wang, Deqing Sun, and Ming-Hsuan Yang,
\newblock ``Drivinggaussian: Composite gaussian splatting for surrounding dynamic autonomous driving scenes,''
\newblock in {\em IEEE Conference on Computer Vision and Pattern Recognition}, 2024, pp. 21634--21643.

\bibitem{yang2023emernerf}
Jiawei Yang, Boris Ivanovic, Or~Litany, Xinshuo Weng, Seung~Wook Kim, Boyi Li, Tong Che, Danfei Xu, Sanja Fidler, Marco Pavone, et~al.,
\newblock ``Emernerf: Emergent spatial-temporal scene decomposition via self-supervision,''
\newblock {\em Arxiv}, 2023.

\bibitem{liu2024one}
Minghua Liu, Chao Xu, Haian Jin, Linghao Chen, Mukund Varma~T, Zexiang Xu, and Hao Su,
\newblock ``One-2-3-45: Any single image to 3d mesh in 45 seconds without per-shape optimization,''
\newblock {\em Neural Information Processing Systems}, vol. 36, 2024.

\bibitem{kerbl20233d}
Bernhard Kerbl, Georgios Kopanas, Thomas Leimk{\"u}hler, and George Drettakis,
\newblock ``3d gaussian splatting for real-time radiance field rendering.,''
\newblock {\em ACM Trans. Graph.}, vol. 42, no. 4, pp. 139--1, 2023.

\bibitem{sabour2024spotlesssplats}
Sara Sabour, Lily Goli, George Kopanas, Mark Matthews, Dmitry Lagun, Leonidas Guibas, Alec Jacobson, David~J Fleet, and Andrea Tagliasacchi,
\newblock ``Spotlesssplats: Ignoring distractors in 3d gaussian splatting,''
\newblock {\em Arxiv}, 2024.

\bibitem{lin2024thermalnerf}
Yvette~Y Lin, Xin-Yi Pan, Sara Fridovich-Keil, and Gordon Wetzstein,
\newblock ``Thermalnerf: Thermal radiance fields,''
\newblock in {\em 2024 IEEE International Conference on Computational Photography (ICCP)}. IEEE, 2024, pp. 1--12.

\bibitem{lu2024thermalgaussian}
Rongfeng Lu, Hangyu Chen, Zunjie Zhu, Yuhang Qin, Ming Lu, Le~Zhang, Chenggang Yan, and Anke Xue,
\newblock ``Thermalgaussian: Thermal 3d gaussian splatting,''
\newblock {\em Arxiv}, 2024.

\bibitem{chen2024thermal3d}
Qian Chen, Shihao Shu, and Xiangzhi Bai,
\newblock ``Thermal3d-gs: Physics-induced 3d gaussians for thermal infrared novel-view synthesis,''
\newblock in {\em European Conference on Computer Vision}. Springer, 2024, pp. 253--269.

\bibitem{ye2024thermal}
Tianxiang Ye, Qi~Wu, Junyuan Deng, Guoqing Liu, Liu Liu, Songpengcheng Xia, Liang Pang, Wenxian Yu, and Ling Pei,
\newblock ``Thermal-nerf: Neural radiance fields from an infrared camera,''
\newblock {\em Arxiv}, 2024.

\bibitem{schonberger2016structure}
Johannes~L Schonberger and Jan-Michael Frahm,
\newblock ``Structure-from-motion revisited,''
\newblock in {\em IEEE Conference on Computer Vision and Pattern Recognition}, 2016, pp. 4104--4113.

\bibitem{chaurasia2013depth}
Gaurav Chaurasia, Sylvain Duchene, Olga Sorkine-Hornung, and George Drettakis,
\newblock ``Depth synthesis and local warps for plausible image-based navigation,''
\newblock {\em ACM Transactions on Graphics (Proceedings of SIGGRAPH)}, vol. 32, no. 3, pp. 1--12, 2013.

\bibitem{mildenhall2020nerf}
Ben Mildenhall, Pratul~P Srinivasan, Matthew Tancik, Jonathan~T Barron, Ravi Ramamoorthi, and Ren Ng,
\newblock ``Nerf: representing scenes as neural radiance fields for view synthesis (2020),''
\newblock {\em Arxiv}, 2020.

\bibitem{yu2021plenoxels}
Alex Yu, Sara Fridovich-Keil, Matthew Tancik, Qinhong Chen, Benjamin Recht, and Angjoo Kanazawa,
\newblock ``Plenoxels: Radiance fields without neural networks,''
\newblock {\em Arxiv}, vol. 2, no. 3, pp. 6, 2021.

\bibitem{muller2022instant}
Thomas M{\"u}ller, Alex Evans, Christoph Schied, and Alexander Keller,
\newblock ``Instant neural graphics primitives with a multiresolution hash encoding,''
\newblock {\em ACM transactions on graphics (TOG)}, vol. 41, no. 4, pp. 1--15, 2022.

\bibitem{wang2004image}
Zhou Wang, Alan~C Bovik, Hamid~R Sheikh, and Eero~P Simoncelli,
\newblock ``Image quality assessment: from error visibility to structural similarity,''
\newblock {\em IEEE Transactions on Image Processing}, vol. 13, no. 4, pp. 600--612, 2004.

\bibitem{yang2024deformable}
Ziyi Yang, Xinyu Gao, Wen Zhou, Shaohui Jiao, Yuqing Zhang, and Xiaogang Jin,
\newblock ``Deformable 3d gaussians for high-fidelity monocular dynamic scene reconstruction,''
\newblock in {\em IEEE Conference on Computer Vision and Pattern Recognition}, 2024, pp. 20331--20341.

\bibitem{lu20243d}
Zhicheng Lu, Xiang Guo, Le~Hui, Tianrui Chen, Min Yang, Xiao Tang, Feng Zhu, and Yuchao Dai,
\newblock ``3d geometry-aware deformable gaussian splatting for dynamic view synthesis,''
\newblock in {\em IEEE Conference on Computer Vision and Pattern Recognition}, 2024, pp. 8900--8910.

\bibitem{ren2024nerf}
Weining Ren, Zihan Zhu, Boyang Sun, Jiaqi Chen, Marc Pollefeys, and Songyou Peng,
\newblock ``Nerf on-the-go: Exploiting uncertainty for distractor-free nerfs in the wild,''
\newblock in {\em IEEE Conference on Computer Vision and Pattern Recognition}, 2024, pp. 8931--8940.

\bibitem{zhang2018unreasonable}
Richard Zhang, Phillip Isola, Alexei~A Efros, Eli Shechtman, and Oliver Wang,
\newblock ``The unreasonable effectiveness of deep features as a perceptual metric,''
\newblock in {\em IEEE Conference on Computer Vision and Pattern Recognition}, 2018, pp. 586--595.

\end{thebibliography}

\end{document}